\documentclass[10pt,twocolumn,letterpaper,dvipsnames]{article}

\usepackage{wacv}
\usepackage{times}
\usepackage{epsfig}
\usepackage{graphicx}
\usepackage{amsmath}
\usepackage{amssymb}

\usepackage[accsupp]{axessibility}
\usepackage{xspace}
\usepackage{ dsfont }
\usepackage{booktabs}
\usepackage{multicol}
\usepackage{multirow}
\usepackage[dvipsnames]{xcolor}
\usepackage[all]{nowidow}
\usepackage{url}

\newcommand{\Lcal}{\mathcal{L}}
\newcommand{\ddetr}{Def\onedot~DETR\xspace}
\DeclareMathOperator{\softmax}{softmax}

\def\wacvPaperID{604} 

\wacvalgorithmstrack   

\wacvfinalcopy 

\ifwacvfinal
\fi

\ifwacvfinal
\usepackage[breaklinks=true,bookmarks=false]{hyperref}
\else
\usepackage[pagebackref=true,breaklinks=true,colorlinks,bookmarks=false]{hyperref}
\fi

\usepackage[noabbrev, capitalise]{cleveref}
\crefformat{footnote}{#2\footnotemark[#1]#3}

\allowdisplaybreaks

\begin{document}

\title{Towards Few-Annotation Learning for Object Detection: \\
Are Transformer-based Models More Efficient ?}


\author{
Quentin Bouniot\footnote[1]{} \footnote[2]{} \and Angélique Loesch\footnote[1]{} \and Romaric Audigier\footnote[1]{} \and Amaury Habrard\footnote[2]{} \footnote[3]{} \\
\footnote[1]{}  Université Paris-Saclay, CEA, LIST, F-91120, Palaiseau, France \\
{\tt\small \{firstname.lastname\}@cea.fr} \\
\footnote[2]{}  Université de Lyon, UJM-Saint-Etienne, CNRS, IOGS,  \\
Laboratoire Hubert Curien UMR 5516, F-42023, Saint-Etienne, France \\
{\tt\small \{firstname.lastname\}@univ-st-etienne.fr} \\
\footnote[3]{} Institut Universitaire de France (IUF)
}

\maketitle

\linepenalty=1000

\begin{abstract}
  For specialized and dense downstream tasks such as object detection, labeling data requires expertise and can be very expensive, making few-shot and semi-supervised models much more attractive alternatives. While in the few-shot setup we observe that transformer-based object detectors perform better than convolution-based two-stage models for a similar amount of parameters, they are not as effective when used with recent approaches in the semi-supervised setting. In this paper, we propose a semi-supervised method tailored for the current state-of-the-art object detector Deformable DETR in the few-annotation learning setup using a student-teacher architecture, which avoids relying on a sensitive post-processing of the pseudo-labels generated by the teacher model. We evaluate our method on the semi-supervised object detection benchmarks COCO and Pascal VOC, and it outperforms previous methods, especially when annotations are scarce. We believe that our contributions open new possibilities to adapt similar object detection methods in this setup as well.

\end{abstract}


\begin{figure}
        \centering
        \includegraphics[width=0.9\linewidth]{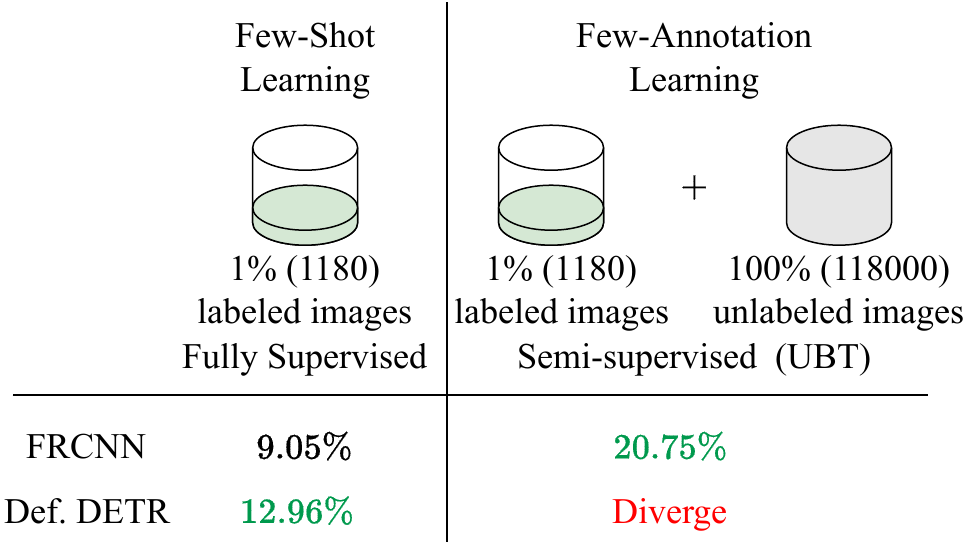}
        \caption{Comparison of mean final performance (mAP in \%) between Faster R-CNN (FRCNN)~\cite{ren2015faster} and Deformable DETR (\ddetr)~\cite{zhu2020deformable} in the Few-Shot and Few-Annotation Learning settings, using only 1\% of labeled data on COCO (about 1180 images). See \cref{sec:exp_details} for experimental details. 
        In the fully supervised case, \ddetr achieves better results than FRCNN. However, in the semi-supervised case implemented in Unbiased Teacher (UBT)~\cite{liu2020unbiased}, \ddetr cannot converge. 
        }
        \label{fig:fal}
\end{figure}

\section{Introduction}

        Deep learning methods are highly successful when trained on a huge amount of \emph{labeled} data. While gathering data is not difficult in most cases, its labeling is always time-consuming and costly. For instance, labeling medical images requires having access to expert knowledge, while annotating images for dense tasks, like object detection and segmentation in autonomous driving, requires going through a tedious process of drawing polygons or bounding boxes around the objects of interest. A more attractive alternative to this process is considered in our work: to guide the learning using only a handful of labeled examples, while simultaneously leveraging a large amount of unlabeled data. This corresponds to a particular case of \emph{semi-supervised learning} (SSL) called \emph{few-annotation learning} (FAL) hereafter.
        
        For the task of \emph{Object Detection} (OD), 
        methods in the literature that tackle this setting~\cite{jeong2019consistency,sohn2020simple,liu2020unbiased,xu2021end,zhou2021instant,tang2021humble} have all considered object detectors based on traditional convolutional networks~\cite{ren2015faster} with a set of specific post-processing heuristics required for them to work~\cite{hosang2017learning,zhang2020bridging}. More recent object detectors are based on an encoder-decoder architecture using transformers \cite{vaswani2017attention} that allows for end-to-end OD without depending on this hand-crafted pipeline~\cite{carion2020end,zhu2020deformable}. However, they have not yet been tested in the SSL context. 
        
        The starting point of this paper is the observation that current state-of-the-art transformer-based architecture~\cite{zhu2020deformable} performs much better than traditional object detectors in a data-scarce fully supervised learning setting, also called \emph{Few-Shot Learning} (FSL), for an equal number of parameters.
        However, when plugging it into a state-of-the-art \emph{Semi-Supervised Object Detection} (SSOD) method~\cite{liu2020unbiased}, we observe that the model fails to converge, meaning that when used as is (Figure~\ref{fig:fal}), applying SSL methods from literature to transformer-based object detectors does not guarantee good results. 
        Thus, we propose a novel SSL method tailored for transformer-based architectures in order to take advantage of the effectiveness of transformers in FSL, and upscale these methods for FAL. Our proposed method achieves state-of-the-art in several FAL benchmarks. 
        
        More precisely, our contributions are summarized as: \\
        1) After showcasing the strong performance of transformer-based detectors using few labeled data, we propose \emph{Momentum Teaching DETR} (MT-DETR), an approach for SSOD that leverages the specificities of transformer-based architectures and outperforms previous semi-supervised approaches in FAL settings.
        2) Contrary to convolution-based OD methods, our approach does not rely on heuristics and post-processing for constructing pseudo-labels. Thus, it eliminates sensitive hyperparameters.



\section{Related Work}\label{sec:related_works}


\subsection{Fully Supervised Object Detection}

Object Detection is a significant and widely studied problem in computer vision~\cite{girshick2014rich,girshick2015fast,ren2015faster, redmon2016you, liu2016ssd, lin2017feature, tian2019fcos}. Essentially, it combines the tasks of object localization and classification. It is a dense task that requires a precise understanding of the image, the objects and their context. The most popular OD models have been based on fully convolutional neural networks~\cite{girshick2014rich, ren2015faster, redmon2016you}. These methods can be separated into \emph{two-stage}~\cite{girshick2014rich,girshick2015fast,ren2015faster,lin2017feature} or \emph{one-stage}~\cite{redmon2016you, liu2016ssd, tian2019fcos} detectors. The former methods make predictions of boxes and their class labels based on region proposals, \eg from a Region Proposal Network (RPN)~\cite{ren2015faster}, while the latter make predictions \wrt to anchors~\cite{lin2017focal} or a grid of possible object centers~\cite{redmon2016you, zhou2019objects, tian2019fcos}.
Their performance depends heavily on hand-designed heuristics, with the most prominent example being the Non-Maximal Suppression (NMS) post-processing, widely used in state-of-the-art OD methods~\cite{hosang2017learning, bodla2017soft}. 
More recently, a novel detector based on an encoder-decoder architecture using transformers~\cite{vaswani2017attention} has been proposed~\cite{carion2020end}. This allows end-to-end detection with a simpler pipeline and eliminates the need for the above-mentioned heuristics. 
The training complexity of this architecture was subsequently improved in Deformable DETR (\ddetr)~\cite{zhu2020deformable}, by changing the attention operations into deformable attention, which leads to an improved convergence speed. 
In this work, we found that \ddetr is a stronger baseline for FSL than the more popular Faster-RCNN~\cite{ren2015faster} widely used in previous work, which motivated us to focus on transformer-based OD architecture.



\begin{table*}[ht]
    \centering
    \resizebox{0.9\linewidth}{!}{%
    \begin{tabular}{@{}llcccccc@{}}
    \toprule
    \multirow{2}{*}{Method} & \multirow{2}{*}{Params.} & \multicolumn{4}{c}{COCO} & \multicolumn{2}{c}{VOC07} \\
    \cmidrule(lr){3-6}
    \cmidrule(lr){7-8}
     & & 0.5\% (590) & 1\% (1180) & 5\% (5900) & 10\% (11800)  & 5\% (250) & 10\% (500) \\
    \midrule
        FRCNN + FPN$^\dagger$ & 42M & $6.83 \pm 0.15$ & $9.05 \pm 0.16$ & $18.47 \pm 0.22$ & $23.86 \pm 0.81$ & $18.47 \pm 0.39$ & $25.23 \pm 0.22$ \\
        \ddetr & 40M & $\textbf{8.95} \pm \textbf{0.51}$ & $\textbf{12.96} \pm \textbf{0.08}$ & $\textbf{23.59} \pm \textbf{0.21}$ & $\textbf{28.55} \pm \textbf{0.08}$ & $\textbf{22.87} \pm \textbf{0.38}$ & $\textbf{29.03} \pm \textbf{0.46}$ \\
        $\Delta$ & & \textcolor{ForestGreen}{$+2.12$} & \textcolor{ForestGreen}{$+3.91$} & \textcolor{ForestGreen}{$+5.12$} & \textcolor{ForestGreen}{$+4.69$} & \textcolor{ForestGreen}{$+4.40$} & \textcolor{ForestGreen}{$+3.80$} \\
    \bottomrule
    \end{tabular}%
    }
    \begin{center}
        \caption{Performance (mAP in \%) comparison between Faster-RCNN (FRCNN)~\cite{ren2015faster} with Feature Pyramid Network (FPN)~\cite{lin2017feature}, a two-stage detector commonly used in SSOD methods, and Deformable DETR (\ddetr)~\cite{zhu2020deformable}, a state-of-the-art transformer-based object detector, with the same ResNet-50 backbone model. 
        The performances are reported for different percentages (and the corresponding number of images) of COCO and VOC07 labeled training data. See \cref{sec:exp_details} for more details on the experiments.
        \ddetr performs better than FRCNN + FPN with fewer labeled data for a similar amount of parameters.
        $^\dagger$: Results from \cite{liu2020unbiased} if available, from our reproduction otherwise.}
        \label{tab:comp_ddetr_frcnn}
    \end{center}
\end{table*}

\subsection{Semi-supervised Learning}
The goal of semi-supervised learning is to take advantage of unlabeled data along with labeled data during training. In the more specific case of FAL, it allows reducing the need of a large amount of labeled data by leveraging the use of unlabeled data.

\textbf{Image Classification}
The problem of SSL in computer vision was historically tackled first for the image classification task, with significant progress made using deep neural networks~\cite{sajjadi2016regularization,tarvainen2017mean,miyato2018virtual,berthelot2019mixmatch,sohn2020fixmatch}. A popular type of approach in this field uses \emph{pseudo-labeling}~\cite{lee2013pseudo,berthelot2019mixmatch,arazo2020pseudo,sohn2020fixmatch}, by generating pseudo-labels from class \emph{predictions} for unlabeled data, either offline~\cite{lee2013pseudo} or online~\cite{berthelot2019mixmatch, sohn2020fixmatch}, and then training on a mix of ground truth and pseudo-labels. Another similar branch of methods is using \emph{consistency regularization}~\cite{sajjadi2016regularization, tarvainen2017mean,LaineA17,chen2020big} to match the predicted class \emph{distributions} of the online version of the model called \emph{student}, to the predicted distributions of a different version of the model called \emph{teacher}, both seeing two \emph{different augmented views} of the inputs.
Following recent trends~\cite{sohn2020fixmatch, chen2020big}, our work takes inspiration from both groups of methods adapted to OD, by training a student model to match the predicted \emph{probability distributions} of proposals made by a teacher model.

\textbf{Object Detection}
Methods in the literature are mainly relying on pseudo-labels provided by a teacher model after applying strong data augmentations on unlabeled data~\cite{jeong2019consistency,sohn2020simple,liu2020unbiased,xu2021end,zhou2021instant,tang2021humble}. 
The use of geometric transformations in these strong augmentations is particularly important for OD~\cite{sohn2020simple}, due to the localization task intrinsic to the problem.  
The most recent and best performing ones~\cite{liu2020unbiased,xu2021end,tang2021humble} are also updating the teacher through Exponential Moving Average (EMA)~\cite{lillicrap2015continuous} of the student's weights to continuously improve the teacher and, thus, the pseudo-labels given to the student.
Although the use of EMA has improved the performance of the models, we propose in our work to stabilize the teacher, by applying an updating strategy throughout training, inspired by recent advances in self-supervised learning~\cite{grill2020bootstrap,caron2021emerging}.
Pseudo-labels are obtained, either by using a \emph{hard labeling}~\cite{jeong2019consistency,sohn2020simple,liu2020unbiased,xu2021end,zhou2021instant} approach, which consists in applying an $\arg \max$ to the predictions, or a \emph{soft labeling}~\cite{tang2021humble} approach, by fully using the predicted distribution. All the previous methods are relying on NMS and thresholding the \emph{confidence scores}, \ie the $\softmax$ of the predictions, given by the teacher model. However, the above-mentioned post-processing steps are sensitive to hyperparameters and introduce a bias into the model incentivizing it to be highly confident in its predictions, which may be suboptimal, particularly when few labeled data are available. Therefore, we aim to remove all these post-processing steps in this work.
Furthermore, SSOD methods in the literature have been exclusively built and evaluated using two-stage OD architectures, and we found that they do not work as is for the more recent detection models based on transformers. 


In this paper, we investigate SSOD through the lens of FAL, and we focus our experiments in this setting, in contrast to previous work that address FAL with only a limited number of experiments.

\section{A semi-supervised learning approach for transformer-based object detection}\label{sec:mt_detr}

In this section, we first motivate our main idea to use a recent state-of-the-art transformer-based OD method in an SSL context by providing several results on both FSL and FAL settings. Then, we present Momentum Teaching DETR (MT-DETR), our transformer-based SSOD method more adapted to FAL and illustrated in Figure~\ref{fig:method}. More specifically, we describe the construction of the pseudo-labels for unlabeled data, and the update scheduling for the teacher model.

\subsection{How do object detectors handle data scarcity ?}

From the results presented in \cref{tab:comp_ddetr_frcnn}, we can see that Deformable DETR (\ddetr) \cite{zhu2020deformable}, a recent state-of-the-art detection model based on transformers, achieves consistently better performance than the most popular two-stage method in FSL. We refer the reader to \cref{sec:exp_details} for all the implementation details. 

These results motivated us to implement \ddetr in a state-of-the-art SSOD method to see how it performs in FAL settings. We opted for the recent Unbiased Teacher (UBT)~\cite{liu2020unbiased}, as its strong results in FAL were easily reproducible with the provided code.
Surprisingly, we observed that with \ddetr detector, the model does not converge in all the FAL settings tested: 1\% of COCO as labeled data (\ie about 1180 labeled images), 5\% and 10\% of VOC 07 (\ie 250 and 500 labeled images respectively). Even though it passes by an early best (about 17\% mAP on 1\% of COCO) at the beginning of training, the model collapses soon after. This diverging behavior is not satisfying in practice, even more so that the same method used with a Faster-RCNN~\cite{ren2015faster} architecture converges (it achieves about 20\% final mAP on 1\% of COCO) in similar settings (\cf \cref{fig:fal}).
All of this shows that current state-of-the-art SSOD methods are not adapted to more recent transformer-based architectures. 

Inspired by these results, we propose an SSL method tailored for transformer-based OD called \emph{Momentum Teaching DETR (MT-DETR)}.


\subsection{Overview of our approach}

\begin{figure*}[!t]
        \centering
        \includegraphics[width=0.9\linewidth]{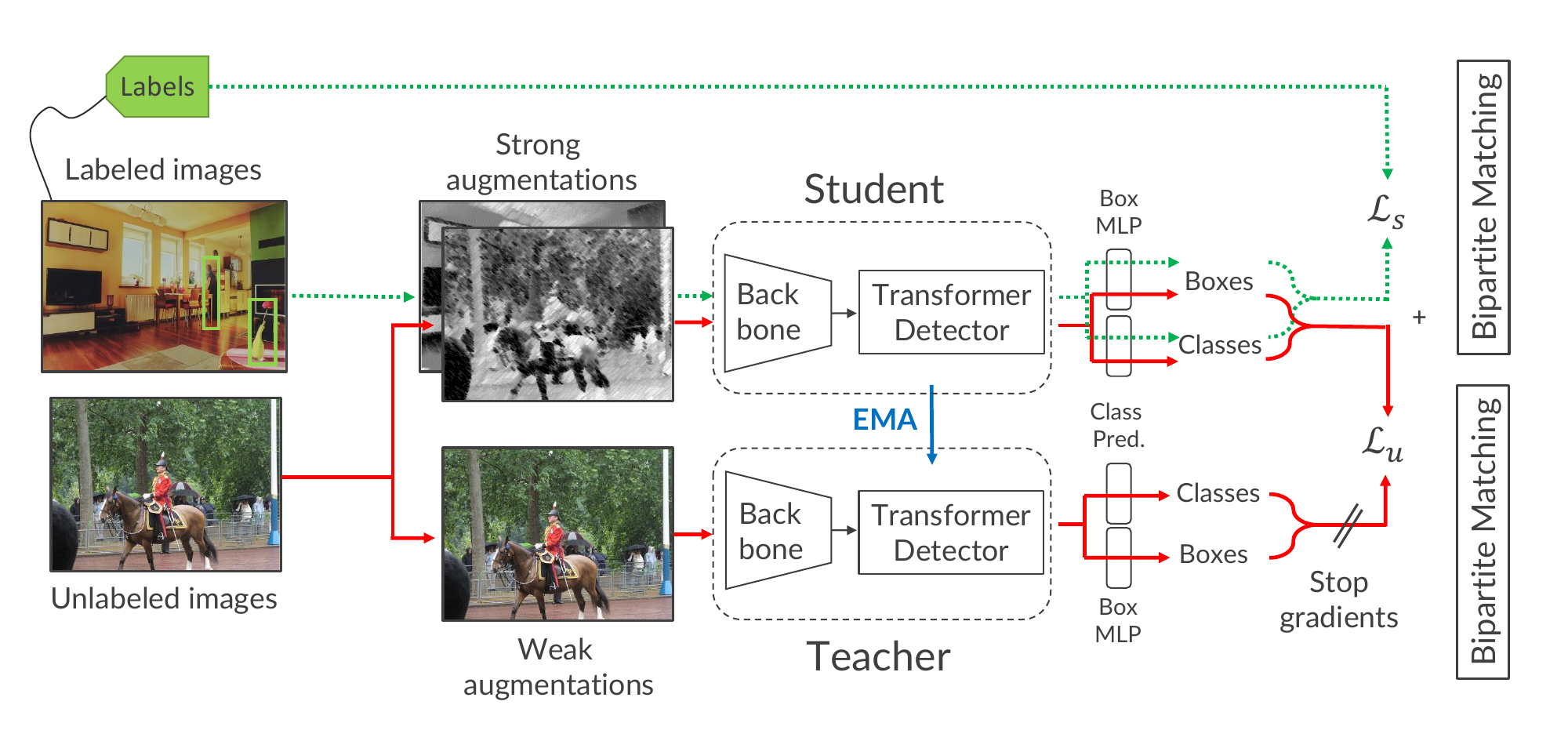}
        \caption{Overview of our Momentum Teaching DETR (MT-DETR) approach for SSOD. The method follows a student-teacher architecture, with the teacher updated through an Exponential Moving Average (EMA) of the student. The keep rate parameter for the EMA follows a \emph{cosine scheduling}. In the supervised branch (\emph{in dotted and \textcolor{ForestGreen}{green}}), the supervised loss $\Lcal_s$ is computed with the predictions of the student on the labeled images. In the unsupervised branch (\emph{in straight and \textcolor{red}{red}}), the \emph{raw}, \ie \emph{unprocessed}, outputs of the teacher model for the weakly augmented unlabeled images are used as \emph{soft} pseudo-labels without applying any heuristic like NMS or confidence thresholding. After finding the best corresponding detection proposals with bipartite matching, the student model learns from the strongly augmented images to match the distribution of class probabilities and the bounding boxes in these pseudo-labels through the unsupervised loss $\Lcal_u$.}
        \label{fig:method}
\end{figure*}
As shown in Figure~\ref{fig:method}, our approach is composed of a \emph{student-teacher architecture}, which is common for semi-supervised learning~\cite{tarvainen2017mean,sohn2020fixmatch}. Both student and teacher models are initialized from a fully supervised model trained on the \emph{few labeled data} available. Then, during the semi-supervised training, the method takes as inputs a batch of labeled images $\mathcal{B}^l = \{(x^l_i, y^l_i)\}_{i=1}^{N^l}$ and a batch of unlabeled images $\mathcal{B}^u = \{x^u_i\}_{i=1}^{N^u}$. We define $x^l_i$ and $x^u_i$ as the $i$\textsuperscript{th} labeled and unlabeled image respectively, $y^l_i = \{y^l_{(i,j)} \}_{j=1}^{k_i} = \{ (c^l_{(i,j)}, b^l_{(i,j)}) \}_{j=1}^{k_i} \in \{ \{1,2,\dots,C\} \times \mathbb{R}^4 \}_{j=1}^{k_i}$ as the corresponding $k_i$ ground truth class labels and box coordinates, and finally, $N^l$ and $N^u$ are respectively the labeled and unlabeled batch sizes.
The student model is updated by a weighted combination of a supervised loss $\Lcal_s$ and an unsupervised loss $\Lcal_u$ with weight $\lambda_u \in \mathbb{R}$:
\begin{equation}
\Lcal(\mathcal{B}^l, \mathcal{B}^u) = \frac{1}{N^l} \Lcal_s(\mathcal{B}^l) + \frac{\lambda_u}{N^u} \Lcal_u(\mathcal{B}^u).
\end{equation}

\noindent Below, we first describe the \emph{supervised branch}, which computes the \emph{supervised loss} using the batch of labeled data $\mathcal{B}^l$. Then, we detail the \emph{unsupervised branch}, which computes the \emph{unsupervised loss} with the batch of unlabeled data $\mathcal{B}^u$. 

\textbf{Supervised branch}
To compute the supervised loss, the supervised branch follows the supervised learning of \ddetr~\cite{zhu2020deformable}, which is an improved version of  DETR~\cite{carion2020end}.
For each image $x^l_i$, the student model infers $N$ predictions $\hat{y}^l_i = \{\hat{y}^l_{(i,j)}\}_{j=1}^N = \{ (\hat{c}^l_{(i,j)},\hat{b}^l_{(i,j)}) \}_{j=1}^N$ of boxes $\hat{b}^l_{(i,j)}$ 
and their associated predicted labels \emph{logits} $\hat{c}^l_{(i,j)} \in \mathbb{R}^{C+1}$, with the ($C+1$)\textsuperscript{th} logit representing the \emph{no object} ($\varnothing$) class. 
Then, the Hungarian algorithm finds from all the permutations of $N$ elements $\mathfrak{S}_N$, the optimal bipartite matching $\hat{\sigma}^l_i$ between the predictions $\hat{y}^l_i$ of the student model and the ground truth labels $y^l_i$:
$\hat{\sigma}^l_i = \arg \min_{\sigma \in \mathfrak{S}_N}  \sum_j^N \Lcal_{\text{match}}(y^l_{(i,j)}, \hat{y}^l_{(i,\sigma(j))}).$
Thus, for each labeled image $x^l_i$, the $j$\textsuperscript{th} ground truth $y^l_{(i,j)}$ is associated to $\hat{\sigma}^l_i(j)$.
Similarly to the loss used in object detectors, the matching cost $\Lcal_{\text{match}}$ used in the Hungarian algorithm takes into account both class and bounding box predictions through a linear combination of the \emph{Focal loss}~\cite{lin2017focal} $\Lcal_{\text{focal}}$, the $\ell_1$ loss of the box coordinates, and the generalized IoU loss~\cite{rezatofighi2019generalized} $\Lcal_\text{giou}$, respectively. These loss functions are then used to compute the supervised loss $\Lcal_s$ as well:
\begin{align}
\begin{split}
& \Lcal_\text{match}(y^l_{(i,j)}, \hat{y}^l_{(i,\sigma(j))}) = \mathds{1}_{\{ \hat{c}^l_{(i,\sigma(j))} \neq \varnothing \}} \Bigl[ \\
& \qquad \lambda_\text{class} \Lcal_\text{focal}\left(c^l_{(i,j)}, \hat{c}^l_{(i,\sigma(j))}\right) \\ 
& \qquad + \lambda_{\ell_1} \|b^l_{(i,j)} - \hat{b}^l_{(i,\sigma(j))}\|_1 \\
& \qquad + \lambda_\text{giou} \Lcal_\text{giou}\left(b^l_{(i,j)}, \hat{b}^l_{(i,\sigma(j))}\right)\Bigr], \\
\end{split} \\
\begin{split}
& \Lcal_s(\mathcal{B}^l) = \sum_{i=1}^{N^l} \sum_{j=1}^N \Bigl[ \lambda_\text{class} \Lcal_{\text{focal}}\left(c^l_{(i,j)}, \hat{c}^l_{(i,\hat{\sigma}^l_i(j))}\right) \\
&\qquad + \mathds{1}_{\{ \hat{c}^l_{(i,\hat{\sigma}^l_i(j))} \neq \varnothing \}} \lambda_{\ell_1} \|b^l_{(i,j)} - \hat{b}^l_{(i,\hat{\sigma}^l_i(j))}\|_1 \\
&\qquad + \mathds{1}_{\{ \hat{c}^l_{(i,\hat{\sigma}^l_i(j))} \neq \varnothing \}} \lambda_\text{giou} \Lcal_{\text{giou}}\left(b^l_{(i,j)}, \hat{b}^l_{(i,\hat{\sigma}^l_i(j))}\right) \Bigr]. 
\end{split}
\end{align}%
\noindent In the above equations, we define $\lambda_\text{class}, \lambda_{\ell_1}, \lambda_\text{giou} \in \mathbb{R}$ as the coefficients in the matching cost and $\mathds{1}_{\mathcal{X}}$ the \emph{indicator function}, such that $\forall x, \mathds{1}_{\mathcal{X}}(x) = 1$ iff $x \in \mathcal{X}$.


\textbf{Unsupervised branch}
Our main contribution is the unsupervised loss for transformer-based OD. In the unsupervised branch, we produce two different views for each unlabeled image $x^u_i$: a \emph{weakly augmented view} ${x^u_i}^\prime$ and a \emph{strongly augmented view} ${x^u_i}''$\footnote{The weak and strong augmentations are described in \cref{sec:exp_details}.}. Then, the teacher model provides \emph{soft pseudo-labels} $y^u_i = \{y^u_{(i,j)}\}_{j=1}^N = \{ (c^u_{(i,j)},b^u_{(i,j)}) \}_{j=1}^N$, with $c^u_{(i,j)}$ the predicted \emph{logits}, for each \emph{weakly augmented} unlabeled image ${x^u_i}^\prime$, and the student model infers predictions $\hat{y}^u_i = \{\hat{y}^u_{(i,j)}\}_{j=1}^N = \{ (\hat{c}^u_{(i,j)},\hat{b}^u_{(i,j)}) \}_{j=1}^N$  from the corresponding \emph{strongly augmented} unlabeled view ${x^u_i}''$.

We apply the same Hungarian algorithm with the same matching cost $\Lcal_{\text{match}}$ to obtain the best permutation $\hat{\sigma}^u_i = \arg \min_{\sigma \in \mathfrak{S}_N}  \sum_j^N \Lcal_{\text{match}}(y^u_{(i,j)}, \hat{y}^u_{(i,\sigma(j))})$, that matches the predictions of the student with the closest pseudo-label.
In the unsupervised loss $\Lcal_u$, we follow the consistency regularization paradigm~\cite{bucilua2006model,chen2020big,caron2021emerging}. We train the student network to match the probability distributions of the classes predicted by the student with the soft pseudo-labels proposed by the teacher. We learn to match these distributions by minimizing the cross-entropy between the two class distribution outputs normalized by a $\softmax$ function. We define respectively:
\begin{equation}
{p^s_{(i,j)}}^{(k)} = \softmax(c^u_{(i,j)})^{(k)} = \frac{\exp({c^u_{(i,j)}}^{(k)})}{\sum_{n=1}^{C+1}\exp({c^u_{(i,j)}}^{(n)})},
\end{equation}

\noindent and ${p^t_{(i,j)}}^{(k)} = \text{softmax}(\hat{c}^u_{(i,j)})^{(k)}$, the student and teacher class distribution outputs, where $c^{(k)}$ is the $k$\textsuperscript{th} element of $c, \forall c \in \mathbb{R}^{C+1}$. Then the cross-entropy loss is defined as:
\begin{equation}
\Lcal_{\text{CE}}(c^u_{(i,j)},\hat{c}^u_{(i,j)})  = - \sum_{k=1}^{C+1} {p^s_{(i,j)}}^{(k)} \log {p^t_{(i,j)}}^{(k)},
\end{equation}
and finally, we compute the unsupervised loss $\Lcal_u$ as:
\begin{equation}
\begin{split}
\Lcal_u(\mathcal{B}^u) = & \sum_{i=1}^{N^u} \sum_{j=1}^N \Bigl[ \lambda_\text{class} \Lcal_{\text{CE}}\left(c^u_{(i,j)}, \hat{c}^u_{(i,\hat{\sigma}^u_i(j))}\right) \\
&+ \mathds{1}_{\{ \hat{c}^u_{(i,\hat{\sigma}^u_i(j))} \neq \varnothing \}} \lambda_{\ell_1} \| b^u_{(i,j)} - \hat{b}^u_{(i,\hat{\sigma}^u_i(j))} \|_1 \\
&+ \mathds{1}_{\{ \hat{c}^u_{(i,\hat{\sigma}^u_i(j))} \neq \varnothing \}} \lambda_\text{giou} \Lcal_{\text{giou}}\left(b^u_{(i,j)}, \hat{b}^u_{(i,\hat{\sigma}^u_i(j))}\right) \Bigr].
\end{split}
\end{equation}

For FAL, we have little information from the labeled data. Therefore, the quality of the pseudo-labels and their contained information play an important part in the training.

\subsection{Construction of the pseudo-labels}

As mentioned above, the unsupervised loss $\Lcal_u$ takes into account the class predictions through a cross-entropy between the outputs of the student model and the matched outputs of the teacher model. We use the $\softmax$ of the outputs of the teacher model as \emph{soft pseudo-labels} for the cross-entropy, as opposed to \emph{hard pseudo-labels} obtained after taking the $\arg \max$.  

Following the DETR philosophy~\cite{carion2020end}, we give to the students the \emph{raw soft pseudo-labels} obtained from the teacher, \ie we remove all handmade heuristics to process the teacher outputs, namely, the NMS and confidence thresholding. Both of these post-processing steps are sensitive to hyperparameters and restrict the diversity in the pseudo-labels. By introducing a bias to keep the most confident proposals, they have the unwanted effect of encouraging the models to always be highly confident in their predictions. 
In the case of FAL, where we have access to only a few labeled examples for each class, the model might not be confident for some classes, leading them to be discarded early by the post-processing. Relying on the model's confidence in certain predictions can be tricky. 
Using the full distributions makes the model less prone to focus on being highly confident in their predictions, and forces the model to take into account the relations between classes. 
Furthermore, the Hungarian algorithm used in transformer-based OD methods leverages the diversity of proposals given by the model and benefits from the fact that the model is not overconfident on a single class thanks to the matching loss. 
Indeed, the bipartite matching can favor proposals with better localizations even if the model is less confident in its class predictions, making the use of raw soft pseudo-labels more suitabled for transformer-based detectors.

To obtain strong and insightful pseudo-labels helping the student, the teacher must be updated throughout training. We describe the update process in the following section.

\subsection{Updating the Teacher model }

To avoid a poor supervision from the teacher, its weights $\theta_t$ are updated by an Exponential Moving Average (EMA) from the student's weights $\theta_s$ using a keep rate $\alpha \in [0,1]$:
\begin{equation}
\theta_t \leftarrow \alpha \theta_t + (1-\alpha) \theta_s.
\end{equation}
For $\alpha=1$, the teacher is constant and for $\alpha=0$ its weights are equal to the student's. Therefore, there is a trade-off between a too high and too low keep rate parameter. 
Inspired by the Self-supervised learning literature~\cite{grill2020bootstrap, caron2021emerging}, we update $\alpha$ following a \emph{cosine scheduling} from $\alpha_{\text{start}}$ to $\alpha_{\text{end}}$:
\begin{equation}
    \alpha \triangleq \alpha_\text{end} - (\alpha_\text{end} - \alpha_{\text{start}}) \cdot (\cos(\pi k /K)+1)/2, 
\end{equation}
with $k$ the current \emph{epoch} and $K$ the maximum number of \emph{epochs}. 
This scheduling stabilizes the teacher model, especially in the last training iterations, to make it converge at the end of training.

\begin{table*}[]
    \centering
    \begin{tabular}{@{}lccccc@{}}
    \toprule
    \multirow{2}{*}{Augmentations} & \multirow{2}{*}{Probability} & \multirow{2}{*}{Parameters} & \multirow{2}{*}{Supervised branch} & \multicolumn{2}{c}{Unsupervised branch} \\
    \cmidrule(lr){5-6}
    & & & & Weak & Strong \\
    \midrule
    Horizontal Flip & 0.5 & -- & \checkmark & \checkmark & \checkmark \\
    \midrule
    Resize & 1.0 & short edge $\in$ range(480,801,32) & \checkmark & \checkmark & \checkmark \\
    \midrule
    \multirow{2}{*}{Color Jitter} & \multirow{2}{*}{0.8} & (brightness, contrast, saturation, hue) & \multirow{2}{*}{\checkmark} & & \multirow{2}{*}{\checkmark} \\
     & & = (0.4, 0.4, 0.4, 0.1) & & & \\
    \midrule
    Grayscale & 0.2 & -- & \checkmark & & \checkmark \\
    \midrule
    Gaussian Blur & 0.5 & $\sigma \in [0.1, 2.0]$ & \checkmark & & \checkmark \\
    \midrule
    \multirow{3}{*}{CutOut} & 0.7 & scale $\in [0.05, 0.2]$, ratio $\in [0.3, 3.3]$ & \checkmark & & \checkmark \\
     & 0.5 & scale $\in [0.02, 0.2]$, ratio $\in [0.1, 6]$ & \checkmark & & \checkmark  \\
     & 0.3 & scale $\in [0.02, 0.2]$, ratio $\in [0.05, 8]$ & \checkmark & & \checkmark \\
    \midrule
    Rotate & 0.3 & degrees $\in [-30,30]$ & & & \checkmark \\
    \midrule
    Shear & 0.3 & $\text{shear}_x \in [-30,30]$, $\text{shear}_y \in [-30,30]$ & & & \checkmark \\
    \midrule
    Rescale + Pad & \multirow{2}{*}{0.5} & $\text{translate}_x \in [0,0.25], \text{translate}_y \in [0,0.25]$ & & & \multirow{2}{*}{\checkmark} \\
    + Translation & & $\text{scale}_x \in [0.25,0.75], \text{scale}_y \in [0.25,0.75]$ & & & \\
    \bottomrule
    \end{tabular}
    \begin{center}
        \caption{The different sets of augmentations used during SSL for each branch. The Horizontal Flip and Resize augmentations follow standard supervised training~\cite{carion2020end,zhu2020deformable}. The Color Jitter, Grayscale, Gaussian Blur and CutOut augmentations follow Unbiased Teacher~\cite{liu2020unbiased} training, and the geometric augmentations (Rotate, Shear, Rescale, Pad and Translation) follow Soft Teacher~\cite{xu2021end} training.}
    \label{tab:augmentations}
    \end{center}
\end{table*}

    

\section{Experimental Results}\label{sec:experiments}

\begin{table*}[]
    \centering
    \resizebox{0.99\linewidth}{!}{%
    \begin{tabular}{@{}llcccc@{}}
    \toprule
    \multirow{2}{*}{Method} & \multirow{2}{*}{OD Arch.} & \multicolumn{4}{c}{COCO} \\
    \cmidrule(lr){3-6}
    & & 0.5\% (590) & 1\% (1180) & 5\% (5900) & 10\% (11800) \\
    \midrule
        STAC~\cite{sohn2020simple} & FRCNN + FPN & $9.78 \pm 0.53$ & $13.97 \pm 0.35$ & $24.38 \pm 0.12$ & $28.64 \pm 0.21$ \\
        Instant-Teaching~\cite{zhou2021instant} & FRCNN + FPN & -- & $18.05 \pm 0.15$ & $26.75 \pm 0.05$ & $30.40 \pm 0.05$ \\
        Humble Teacher~\cite{tang2021humble} & FRCNN + FPN & -- & $16.96 \pm 0.38$ & $27.70 \pm 0.15$ & $31.61 \pm 0.28$ \\
        Unbiased Teacher~\cite{liu2020unbiased} & FRCNN + FPN & $16.94 \pm 0.23$ & $20.75 \pm 0.12$ & $28.27 \pm 0.11$ & $31.50 \pm 0.10$ \\
        Soft Teacher~\cite{xu2021end} & FRCNN + FPN & -- & $20.46 \pm 0.39$ & $30.74 \pm 0.08$ & $34.04 \pm 0.14$ \\
    \midrule
        MT-DETR \emph{(Ours)} & \ddetr & $\textbf{17.84} \pm 0.54$ \textcolor{ForestGreen}{\footnotesize{(+8.89)}} & $\textbf{22.03} \pm 0.17$ \textcolor{ForestGreen}{\footnotesize{(+9.07)}} & $\textbf{31.00} \pm 0.11$ \textcolor{ForestGreen}{\footnotesize{(+7.41)}} & $\textbf{34.52} \pm 0.07$ \textcolor{ForestGreen}{\footnotesize{(+5.97)}} \\
    \bottomrule
    \end{tabular}
    }%
    \begin{center}
        \caption{Performance (mAP in \%) of our proposed approach on FAL-COCO, using different percentage of labeled data (with the corresponding number of images reported) and 100\% of the dataset as unlabeled data. For our method, we also indicate the improvements (in \textcolor{ForestGreen}{green} and in p.p\onedot) \wrt the FSL baseline (\cf \cref{tab:comp_ddetr_frcnn}).}
    \label{tab:perf_coco}
    \end{center}
\end{table*}

\begin{table*}[]
    \centering
    \begin{tabular}{@{}llccc@{}}
    \toprule
    \multirow{2}{*}{Method} & \multirow{2}{*}{OD Arch.} & \multicolumn{3}{c}{VOC 07-12} \\
    \cmidrule(lr){3-5}
    & & 5\% (250) & 10\% (500) & 100\% (5000) \\
    \midrule
        STAC~\cite{sohn2020simple} & FRCNN + FPN & -- & -- & 44.64 \\
        Instant-Teaching~\cite{zhou2021instant} & FRCNN + FPN & -- & -- & 50.00 \\
        Humble Teacher~\cite{tang2021humble} & FRCNN + FPN & -- & -- & 53.04 \\
        Unbiased Teacher [github] & FRCNN + FPN & -- & -- & 54.48 \\
        Unbiased Teacher$^*$ & FRCNN + FPN & $35.98 \pm 0.71$ & $40.34 \pm 0.95$ & 54.61 \\
    \midrule
        MT-DETR \emph{(Ours)} & \ddetr & $\mathbf{36.95} \pm 0.53$ \textcolor{ForestGreen}{\footnotesize{(+14.08)}} & $\mathbf{43.15} \pm 1.10$ \textcolor{ForestGreen}{\footnotesize{(+14.12)}} & \textbf{56.2} \\ 
    \bottomrule
    \end{tabular}
    \begin{center}
        \caption{Performance (mAP in \%) of our proposed approach on VOC with fully labeled VOC07 and unlabeled VOC12 to compare with previous work, and in the novel FAL-VOC 07-12 settings. Different percentage of VOC07 are used as labeled data (5\%, 10\% or 100\%, with the corresponding number of images reported), and the full VOC12 dataset is used as unlabeled data. For our method, we also indicate the improvements (in \textcolor{ForestGreen}{green} and in p.p\onedot) \wrt the FSL baseline (\cf \cref{tab:comp_ddetr_frcnn}). $^*$ indicates our implementation of Unbiased Teacher~\cite{liu2020unbiased} in this novel setting to compare with our approach. [github] : updated results after publication~\cite{liu2020unbiased} taken from their official code released\cref{fn:ubt_code}.}
    \label{tab:perf_voc}
    \end{center}
\end{table*}

In this section, we present a comparative study of the results of our method to the state-of-the-art on FAL benchmarks, as well as an ablative study on the most relevant parts. Before that, we detail the datasets, the evaluation and training settings used for the different experiments. 

\subsection{Datasets, evaluation and training details}\label{sec:exp_details}

\textbf{Datasets and evaluation protocol} 
To evaluate our proposed method, we use the MS-COCO (COCO)~\cite{lin2014microsoft} and Pascal VOC (VOC)~\cite{everingham2010pascal} datasets which are standard for object detection, following the settings of existing works~\cite{jeong2019consistency,sohn2020simple,liu2020unbiased,xu2021end,zhou2021instant,tang2021humble}. COCO is a dataset with 80 classes, and VOC contains 20 classes.
We are specifically interested in two \emph{Few Annotation Learning} (FAL) settings: \\
On \emph{FAL-COCO}, we randomly sample 0.5, 1, 5 or 10\% (respectively about 590, 1180, 5900 and 11800 images) of the training set (\emph{train2017}) used as the labeled set and use the full training set for the unlabeled set (about 118k images). Performance is evaluated on \emph{val2017}. \\
On \emph{FAL-VOC 07-12}, we restrict the labeled training set (VOC07 \emph{trainval}) to a random sample of 5 or 10\% (respectively 250 and 500 labeled images), and use the full VOC12 \emph{trainval} (about 11k images) as unlabeled training set. We introduce this novel setting to evaluate our approach in a FAL setting on VOC. We also compare the results with previous SSOD methods using the full VOC07 \emph{trainval} labeled training set (5k labeled images) and VOC12 \emph{trainval} as unlabeled training set. Performance is evaluated on VOC07 \emph{test} set. 

In all settings, performance is reported and compared using the $AP_{50:95}$ (mAP, in \%) evaluation metric using the official COCO and VOC evaluation codes, respectively.

\textbf{Training}
For a fair comparison, a fully supervised ResNet-50~\cite{he2016deep} pretrained on ImageNet~\cite{deng2009imagenet} is used as a backbone for all the methods. 
For fine-tuning \ddetr~\cite{zhu2020deformable} on the few labeled data, we train the model with a batch size of 32 images on 8 GPUs until the validation performance stops increasing, \ie for COCO, up to 2000 epochs for 1\%, 500 epochs for 5\%, 400 epochs for 10\%, and for Pascal VOC, up to 2000 epochs for both 5\% and 10\%. 
For semi-supervised learning, we train MT-DETR for 50 (respectively, 250) epochs of the unlabeled data on COCO (respectively, Pascal VOC) with a batch size of 48 labeled images and 48 unlabeled images (respectively, 24 and 24) on 8 GPUs. All experiments with less than 100\% of labeled data are reproduced on 3 different random subsets\footnote{\:\href{https://github.com/CEA-LIST/MT-DETR}{See our official repository.}}.
The training hyperparameters, are defined as in \ddetr~\cite{zhu2020deformable}. 
The coefficients for the losses are set as $\lambda_\text{class}=2, \lambda_{\ell_1}=5, \lambda_\text{giou}=2$, and $\lambda_u = 4$. Following the training schedule of \ddetr, we always decay the learning rates by a factor of 0.1 after about 80\% of training. The keep rate parameter $\alpha$ follows a \emph{cosine scheduling} from $\alpha_\text{start} = 0.9996$ to $\alpha_\text{end} = 1$, with the value of $\alpha_\text{start}$ chosen according to previous work~\cite{liu2020unbiased}. \\
When using Unbiased Teacher~\cite{liu2020unbiased}, we follow the official implementation\footnote{\label{fn:ubt_code}\noindent \:\href{https://github.com/facebookresearch/unbiased-teacher}{Official UBT repository.}} and the hyperparameters provided.

\textbf{Augmentations}
For strong and weak data augmentations, we follow the common data augmentations used in previous works~\cite{sohn2020simple,liu2020unbiased,xu2021end}. We apply a random resizing and random horizontal flip for weak augmentations. We randomly add color jittering, grayscale, Gaussian blur, CutOut patches for strong augmentations and also randomly add rescaling, translation with padding, shearing and rotating as geometric transformations~\cite{sohn2020simple} in strong augmentations. 
In the supervised branch, images are also randomly augmented using weak and strong augmentations without any geometric transformations following Soft Teacher~\cite{xu2021end} practices. It helps the student model to be augmentation-agnostic, to better predict pseudo-labels coming from non-augmented images in the unsupervised branch. We remove the CutOut augmentation in the supervised branch in the most difficult settings of FAL-COCO 0.5\% and 1\%, since it can cover the only labeled small boxes available and is counterproductive.  
All the parameters for the different augmentations can be found in Table~\ref{tab:augmentations}.

\begin{table*}[]
    \centering
    \resizebox{0.93\linewidth}{!}{%
    \begin{tabular}{@{}lcccccccccc@{}}
    \toprule
    \multirow{2}{*}{Ablative Variant} & \multicolumn{2}{c}{EMA Scheduling} & \multicolumn{2}{c}{Initialization} & \multirow{2}{*}{\textcolor{red}{NMS}} & \multicolumn{4}{c}{Confidence Thresholding} & \multirow{2}{*}{mAP (in \%)} \\
    \cmidrule(lr){2-3}
    \cmidrule(lr){4-5}
    \cmidrule(lr){7-10}
    & \textcolor{ForestGreen}{\textbf{Cosine}} & \textcolor{red}{Constant} & \textcolor{ForestGreen}{\textbf{After FT}} & \textcolor{red}{From scratch} & & \textcolor{ForestGreen}{\textbf{\o}} & \textcolor{red}{0.5} & \textcolor{red}{0.7} & \textcolor{red}{0.9} & \\
    \midrule
    Best & \checkmark & & \checkmark & & & \checkmark & & & & \textbf{22.25} \\
    \midrule
    Abl\onedot Sched\onedot &  & \checkmark & \checkmark & & & \checkmark & & & & 21.48 \\
    \midrule
    Abl\onedot Init\onedot & \checkmark &  & & \checkmark & & \checkmark & & & & 16.51 \\
    \midrule
    Abl. NMS & \checkmark &  & \checkmark & & \checkmark & \checkmark & & & & 19.85 \\
    \midrule
    \multirow{3}{*}{Abl\onedot Thresh\onedot} & \checkmark &  & \checkmark & & & & \checkmark & & & 10.26 \\
     & \checkmark &  & \checkmark & & & & & \checkmark & & 17.34 \\
     & \checkmark &  & \checkmark & & & & & & \checkmark & 12.37 \\
    \bottomrule
    \end{tabular}%
    }
    \begin{center}
        \caption{Ablation studies of the different parts of our method. \textcolor{ForestGreen}{\textbf{Green and bold columns names}} indicate a \emph{positive} effect on the performance and \textcolor{red}{red columns} a \emph{negative} effect. The use of \emph{cosine scheduling}, \emph{an initialization after fine-tuning (FT)} and \emph{raw soft pseudo labels} corresponds to the best combination found.}
        \label{tab:abl_studies}
    \end{center}
\end{table*}


\subsection{Results of FAL on COCO and Pascal VOC}

Tables~\ref{tab:perf_coco} and~\ref{tab:perf_voc} present the results (mAP in \%) obtained by our method compared to previous methods in the literature on the FAL-COCO and FAL-VOC 07-12 benchmarks. As can be seen in both tables, our approach is the only one to consider a transformer-based OD architecture (\ddetr), as opposed to the commonly used two-stage architecture (FRCNN + FPN). When we implemented \ddetr into Unbiased Teacher~\cite{liu2020unbiased} (UBT), we found that the model cannot converge in FAL settings (\cf \cref{fig:fal}). 

First, we can see from both tables that our method always improve performance over the corresponding fully supervised FSL baseline (\cf \cref{tab:comp_ddetr_frcnn}). 
With our method, we outperform state-of-the-art results on all labeled fractions of the dataset, and obtain even more strong results specifically when the annotations are scarce: globally about +1 performance point (p.p\onedot) when using 1k or less labeled images, which is even more significant when the overall performance is low. For FAL-COCO with 1\% of labeled images, our method achieves a mean of 22.03 mAP, which is about 1.2 p.p\onedot, or 6\% of improvement over the state-of-the-art, UBT. Notably, on FAL-VOC with 10\% of labeled images, we obtain mean performance of 43.15 mAP, corresponding to 2.81 p.p\onedot or 7\% of improvement over UBT. We note that our method also outperform the state-of-the-art when using more labeled data, such as with the 100\% labeled VOC07 setting, where we improve of about 1.5 p.p\onedot over UBT.

\subsection{Ablation studies}
In Table~\ref{tab:abl_studies}, we present an ablation study on the main parts of our approach. We review each ablation below.

\textbf{EMA scheduling}
The effect of the EMA scheduling is compared between the \emph{Best} and \emph{Abl\onedot Sched\onedot} rows. We can see that using a \emph{cosine scheduling} to gradually reduce the EMA keep rate parameter $\alpha$ leads to an improvement of about 0.7 p.p., as opposed to using a \emph{constant} value for $\alpha$ as done in other SSL approaches~\cite{liu2020unbiased,xu2021end,tang2021humble}.

\textbf{Initialization}
In this ablation, we study the effect of \emph{end-to-end semi-supervised learning}~\cite{xu2021end} in the row \emph{Abl\onedot Init\onedot} which consists in starting the semi-supervised training \emph{from scratch} compared to an initialization \emph{after Fine-Tuning (FT)} in the row \emph{Best}, in which we initialize both student and teacher models from the weights of the fine-tuned model on the few labeled data. 
As can be seen in \cref{tab:abl_studies} and contrary to Soft Teacher~\cite{xu2021end}, starting the semi-supervised training from fine-tuned weights is much more effective (about 5.7 p.p\onedot better) than starting from randomly initialized weights, since the teacher model will give useful pseudo-labels to the student from the start of training.  

\textbf{NMS}
The importance of removing NMS to avoid filtering interesting pseudo-labels and introducing bias is showcased between the rows \emph{Best} and \emph{Abl\onedot NMS}. We can see that, contrary to the common practice when using other detectors~\cite{liu2020unbiased,xu2021end,tang2021humble}, the introduction of NMS leads to a performance drop of about 2.5 p.p. This is why we used \emph{raw pseudo labels}, \ie without any post-processing.

\textbf{Confidence Thresholding}
The effect of introducing a threshold to filter out the pseudo-labels given by the teacher with poor confidence is shown in the rows \emph{Best} and \emph{Abl\onedot Thresh\onedot}. We test the results using several common values in the literature (0.5, 0.7 and 0.9)~\cite{sohn2020fixmatch,liu2020unbiased,xu2021end}. A value of 0.7 seems to give the best final results (17.34 mAP) between the thresholding variants, but we can see that choosing the best threshold to apply is extremely sensitive. Similarly to Humble Teacher~\cite{tang2021humble}, we also found that removing the confidence threshold to use all the \emph{soft pseudo-labels}, which corresponds to the column with~\o, leads to stronger results (22.24 mAP), less sensitivity and fewer hyperparameters. 


\section{Conclusion}

In this work, we experimented in different data scarce settings with the state-of-the-art transformer-based object detector \ddetr~\cite{zhu2020deformable} and showed that it performs much better than the most popular two-stage detector Faster-RCNN~\cite{ren2015faster} with FPN~\cite{lin2017feature}. Surprisingly, we found that Unbiased Teacher~\cite{liu2020unbiased}, a state-of-the-art SSOD method, did not converge when applied with \ddetr. 

To address this issue, we propose Momentum Teaching DETR (MT-DETR), an SSL approach tailored for OD based on transformers, in order to leverage their good results with few labeled data.
Our method is based on a student-teacher architecture and, contrary to common practice, discards all previously used handcrafted heuristics to process pseudo-labels generated by the teacher. 
These processing steps are sensitive to hyperparameters, and introduce biases with the unwanted effect of forcing the models to be overconfident in their predictions.
We show that our proposed MT-DETR outperforms state-of-the-art methods, especially in FAL settings. 
Future works could push the data scarcity in OD even further to consider very few labeled examples for each class, and better understand how to match the performance of SSL methods for image classification in this setting~\cite{sohn2020fixmatch}. 


\textbf{Acknowledgements}
The authors would like to thank Ievgen Redko for fruitful discussions and proofreading. This work was made possible by the use of the Factory-AI supercomputer, financially supported by the Ile-de-France Regional Council.

\clearpage

{\small
\bibliographystyle{ieee_fullname}
\bibliography{references}
}







\end{document}